\documentclass[lettersize,journal]{IEEEtran}

\usepackage{graphicx}
\usepackage{cite}
\usepackage{picinpar}
\usepackage{amsmath}
\usepackage{url}
\usepackage[latin1]{inputenc}
\usepackage{colortbl}
\usepackage{soul}
\usepackage{multirow}
\usepackage{pifont}
\usepackage{color}
\usepackage{alltt}
\usepackage{hyperref}
\usepackage{enumerate}
\usepackage{siunitx}
\usepackage{breakurl}
\usepackage{epstopdf}
\usepackage{pbox}
\usepackage[ruled,linesnumbered]{algorithm2e}
\usepackage{booktabs}

\usepackage{bm}
\usepackage{hyperref}
\usepackage{amssymb}
\usepackage{romannum}
\usepackage[caption=false,font=footnotesize]{subfig}
\AtBeginDocument{\pagenumbering{arabic}}

\begin{document}

\title{\LARGE \bf
Learning Cloth Folding Tasks with Refined Flow\\ Based Spatio-Temporal Graphs 
}

\author{Peng Zhou, Omar Zahra, Anqing Duan, Shengzeng Huo, Zeyu Wu and David Navarro-Alarcon
\thanks{
This work is supported in part by the Research Grants Council of Hong Kong under grants 14203917 and 15212721, in part by the Key-Area Research and Development Program of Guangdong Province 2020 under project 76, and in part by the Jiangsu Industrial Technology Research Institute Collaborative Research Program Scheme under grant ZG9V.
}
\thanks{All authors are with The Hong Kong Polytechnic University, Department of Mechanical Engineering, Hung Hom, KLN, Hong Kong.
}
}


\maketitle
\thispagestyle{plain}
\pagestyle{plain}

\begin{abstract}
Cloth folding is a widespread domestic task that is seemingly performed by humans but which is highly challenging for autonomous robots to execute due to the highly deformable nature of textiles; It is hard to engineer and learn  manipulation pipelines to efficiently execute it. 
In this paper, we propose a new solution for robotic cloth folding (using a standard folding board) via learning from demonstrations. Our demonstration video encoding is based on a high-level abstraction, namely, a refined optical flow-based spatiotemporal graph, as opposed to a low-level encoding such as image pixels. By constructing a new spatiotemporal graph with an advanced visual corresponding descriptor, the policy learning can focus on key points and relations with a 3D spatial configuration, which allows to quickly generalize across different environments. To further boost the policy searching, we combine optical flow and static motion saliency maps to discriminate the dominant motions for better handling the system dynamics in real-time, which aligns with the attentional motion mechanism that dominates the human imitation process. 
To validate the proposed approach, we analyze the manual folding procedure and developed a custom-made end-effector to efficiently interact with the folding board. 
Multiple experiments on a real robotic platform were conducted to validate the effectiveness and robustness of the proposed method.
\end{abstract}


\IEEEpeerreviewmaketitle

\newcommand\myworries[1]{\textcolor{red}{#1}}

\section{Introduction}
\IEEEPARstart{R}{obots} have become a crucial part of advancing the manufacturing industry in the past few decades and are widely used in domestic environments nowadays \cite{yin2021modeling}.
Cloth folding is high on the list of monotonous home duties that many people dislike and could, theoretically, be performed by a service robot \cite{miller2012geometric}, which seems to be simple tasks for a human, yet, it poses significant challenges for a robotic system. 
Most industrial robots were designed to do repetitive tasks with rigid objects. Clothes, however, require several additional skills currently unavailable to robots \cite{zhou2021lasesom}. 

In contrast to their rigid counterpart, deformable objects (e.g., fabric) may deform substantially and form wrinkles or folds \cite{navarro2016automatic}, significantly increasing the difficulty of its manipulation. 
\cite{navarro2014visual} and \cite{navarro2018fourier} perform successful manipulation tasks on elastic objects by estimating a deformation model from vision and motion sensory feedback. 
However, creasing of clothes upon force exertion complicates the process of modelling and prediction, and complicates the visual perception of the cloth state as well.
Thus, research efforts \cite{li2014real, li2014recognition, borras2020encoding} have been focused on this complex state representation and estimation.


\begin{figure}[t]
	\centering
	\includegraphics[width=\columnwidth
	]{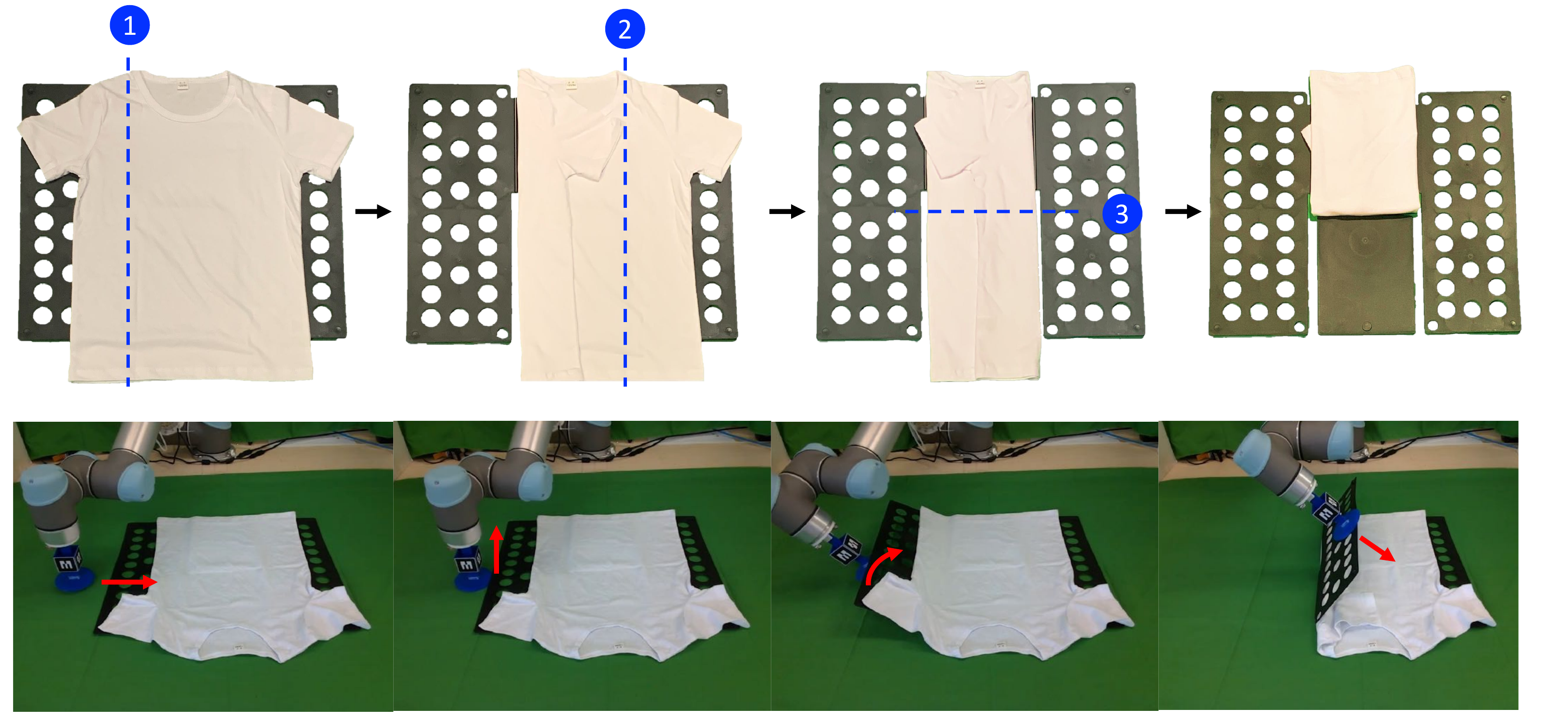}
	\caption{(Up) The folding process with folding board can be divided into \textit{Left folding}, \textit{Right folding} and \textit{Mid-folding}. 
	(Down) Each robotic folding process can be decomposed into  \textit{Approaching}, \textit{Lifting}, \textit{Rotating} and \textit{Pushing} steps.}
	\label{fig:process}
\end{figure}


To tackle these challenges, we propose a new approach to perform the cloth folding task by using a standard manual-folding board, as shown in Fig. $\ref{fig:setup}(\mathrm{c}$).
Instead of designing a complex manipulator \cite{jia2019cloth} or a versatile gripper \cite{donaire2020versatile} to directly have contact with clothes based on real-time measurements on the highly deformable textiles, we aim to solve clothes folding by using this assistive tool to construct limited and unified  manipulation primitives, which allows us to only estimate few intermediate states, instead of measuring deformations frequently.
 Fig. $\ref{fig:process}$ shows a folding process by using a folding board, and the blue dashed lines represent the folding line of each operation. 
 According to the common practice in daily life and benchmarks for cloth manipulation \cite{garcia2020benchmarking}, we divide the folding task into three sub-tasks, namely, \romannum{1}) \textit{left folding}; \romannum{2}) \textit{right folding}; and \romannum{3}) \textit{mid-folding}. Subsequently, each sub-folding could be partitioned into four processes, namely,  \romannum{1}) \textit{approaching}; \romannum{2}) \textit{lifting}; \romannum{3}) \textit{rotating} and \romannum{4}) \textit{pushing} (see details in Sec. \ref{sec:result}).

On the other hand, the ability of Learning from Demonstration (LfD) \cite{kuniyoshi2015learning} - called imitation learning \cite{pathak2018zero} or third-person imitation learning \cite{stadie2017third} - has long been a desirable objective in artificial intelligence as a method to train agents intuitively rather than hard-coding their actions rapidly. For visual imitation to be successful, the demonstrator's sensory environment must be well understood, including how it changes over time \cite{sieb2020graph}. Visual imitation is thus reduced to learning a visual similarity function between the demonstration and imitation scenes, which might be maximized by the imitated behaviors, resulting in accurate skill imitator behavior. This similarity function establishes which elements of the visual observations are essential to replicate the demonstrated behaviors. Put differently, and it determines what to imitate and what to disregard \cite{schaal1999imitation}. Consequently, we formulate the cloth folding task as an imitation learning problem using an assistive folding board. In detail, the robot agent needs to search a policy to reproduce the successful folding task from video sequences of presented demonstrations by a human expert.

\begin{figure*}[tbph]
\centering
\noindent\includegraphics[width=18cm,height=8cm]{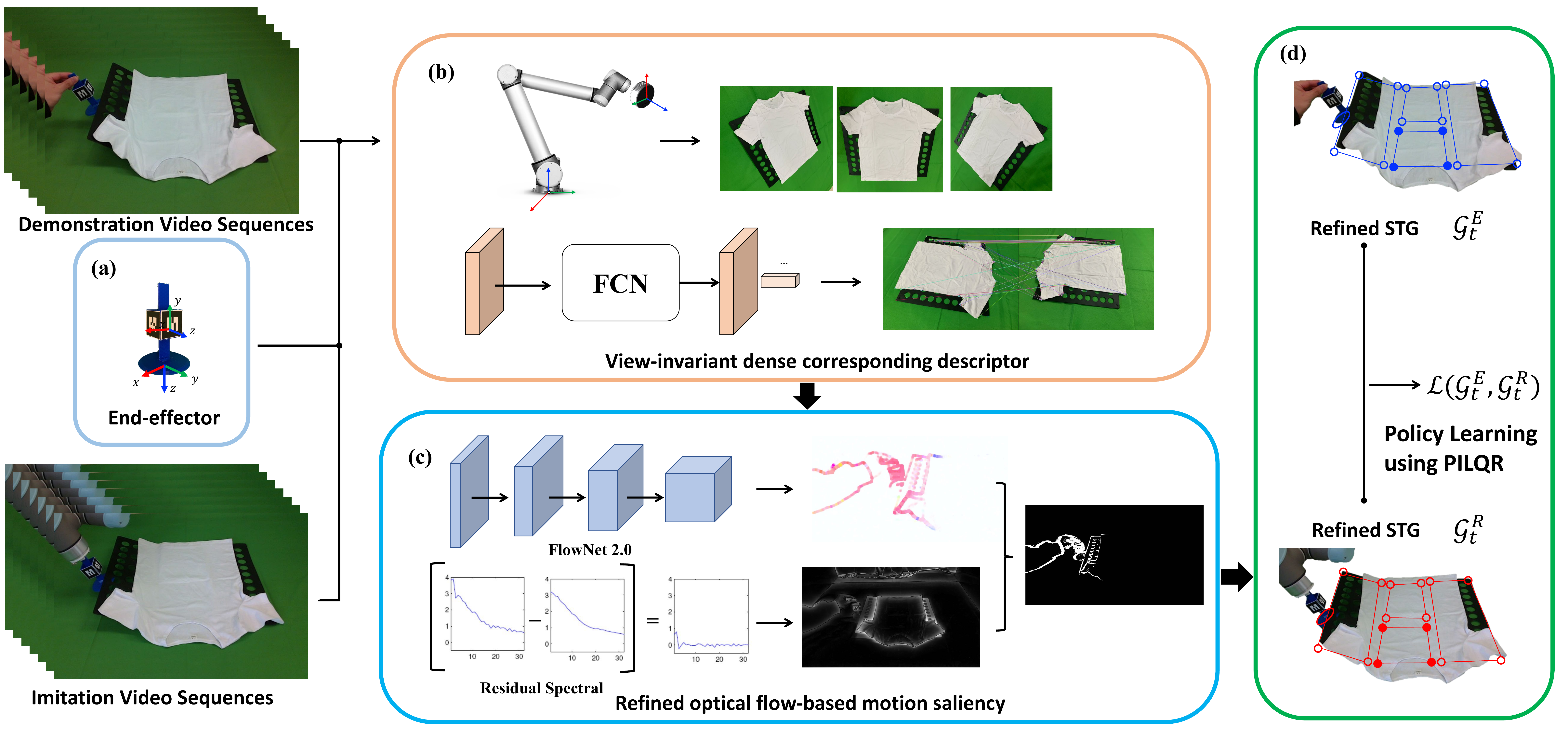}
\caption{Conceptual representation of the proposed approach for cloth folding tasks, which encodes the video sequences into a refined optical flow-based STG across the demonstrator's and imitator's workspace. A well-designed end-effector and a view-invariant dense corresponding descriptor can form the basic structure of an STG, which will be refined with the optical flow-based motion saliency map to optimize the corresponding folding policy.
}
\label{fig:framework}
\end{figure*}

Imitation learning tackles the issue of skill acquisition via demonstration observation \cite{osa2018algorithmic}. However, most prior methods \cite{hussein2017imitation, zhang2018deep} presuppose that demonstrations are provided in the agent's workplace. 
Typically, a mapping between demonstrator and imitator spatial observations is needed and is critical for effective imitation \cite{nehaniv2002imitation}. 
Direct comparison of pixel intensities, on the other hand, is not a reliable indicator of resemblance since it may be tainted by the difference in viewpoints \cite{sharma2019third}, lighting changes \cite{ollis2007bayesian}, or changed poses \cite{zeestraten2017approach}.
Recent method \cite{dwibedi2018learning} has learned such visual similarity by simply training and matching entire image feature embeddings and bypass the specific identification of the environment structure in terms of detected objects' positions and orientations. 
Instead, our method tries to capture a high-level abstraction input first and then train a policy to better handle spatiotermporal dynamics, which is proved to have benefits for generalization of action-conditioned dynamics. \cite{sanchez2018graph} has used such graph-encodings to learn a model predictive control in a real-world application.
To solve imitation learning for cloth folding tasks, we propose a hierarchical encoding for dominant motion representations extracted from a single demonstration video, called a refined optical flow-based spatiotemporal graph (STG), where vertices represent corresponding key points of the manipulated objects (i.e., the folding board) tracked throughout space and time, and edges denote their relative 3D spatial configurations. Our demonstration video encoding is based on a high-level abstraction, a refined optical flow-based STG, instead of a low-level encoding (i.e., image pixels).
For each pair of time steps, we build two refined STGs (see Fig. \ref{fig:framework}d), one for the demonstration workspace and one for the imitation. 
Then, our reward function measures the dissimilarity between corresponding 3D relational vertex pairs and optimizes the policy updated with reinforcement learning algorithms from a single cloth folding demonstration video.
To validate the effectiveness of the proposed approach, we conducted multiple experimental studies on a real robotic platform. 
In summary, the original contributions of this work are as follows:
\begin{itemize}
	\item A new approach for cloth folding task based on an assistive folding board to avoid complex measurements of the highly deformable clothes.
	\item A refined optical flow-based STG designed for detecting dominant motion fields with the aim of accelerating the valid policy searching.
	\item A hierarchical abstraction (i.e., STG) for demonstration video encoding.
\end{itemize}
Although this study involves solving the problem of clothes folding, the proposed methods target a broad set of manipulation problems. Specifically, it targets problems involving manipulation of non-rigid objects through imitation while considering both spatial and temporal graph-structured cues demonstrated by the teaching agent.

The rest of this letter is organized as follows: 
Sec. \ref{sec:formulate} formulates the problem;
Sec. \ref{sec:method} presents the proposed approach; Sec. \ref{sec:result} reports the experiments; Sec. \ref{sec:con} gives final conclusions.
A video of the conducted experiments can be accessed at: \href{https://sites.google.com/view/learnfolding}{\color{blue} https://sites.google.com/view/learnfolding}.


\section{Formulation} \label{sec:formulate}
We define two spatio-temporal graph sequences $ \mathcal{G}_t^{E} = \{\mathcal{V}_t^{E}, \mathcal{E}_t^{E} \mid t = 1, 2, \cdots, N \}$ and $ \mathcal{G}_t^{R} = \{\mathcal{V}_t^{R}, \mathcal{E}_t^{R} \mid t = 1, 2, \cdots, N \}$ extracted from a demonstration video of a human expert and an imitation video of a robotic agent, respectively, in the same time sequence length denoted by $N$. 
This high-level abstraction aligns with our human attentional mechanism during the learning process. 
Therefore, this built graph is independent of its workspace as long as it can be encoded from the observation successfully.
Formally, let $\mathcal{V}_t = \left\{v_t^{1}, \cdots, v_t^{n} \mid v_t^{i} \in \mathbb{R}^{3}\right\}$ be the set of vertex of the corresponding spatio-temporal graph at the $t$ time step and $v_t^i$ represents the $i$-th visual corresponding point during the manipulation process. A vertex $v_t^{i}$ could be an object's position, object part's position, or any interest point detected from the observation during the demonstration or imitation process. 
Subsequently, $\mathcal{E}_t = \left\{e_t^{1}, \cdots, e_t^{m} \mid e_t^{j} \in \{v_t^j, v_t^{j+1}\} \right\} \subseteq \mathcal{V}_{t} \times \mathcal{V}_{t}$ represents the edge set and $e_t^j$ denotes the $j$-th spatial edge to be maintained during a manipulation process between corresponding vertices, $v_t^j$ and $v_t^{j+1}$. 
At the same time step $t$, the graph of the expert's demonstration $\mathcal{G}_t^{E}$ and the one of the robot agent $\mathcal{G}_t^{R}$ have the same structure (see Fig. \ref{fig:framework}(d)), which means same cardinal number for the corresponding set, $\mathbf{card}(\mathcal{V}_t^{E}) = \mathbf{card}(\mathcal{V}_t^{R})$, $\mathbf{card}(\mathcal{E}_t^{E}) = \mathbf{card}(\mathcal{E}_t^{R})$
and same vertex pair for any edge, $ e_t^{E,j} = e_t^{R,j} \mid \forall e_t^{E,j} \in \mathcal{V}_t^E, e_t^{R,j} \in \mathcal{V}_t^{R} $. During the entire manipulation, a graph vertex $v_t$ and a graph edge $e_t$ could be added into or deleted from the spatio-temporal graph with the manipulated object changing. We only ensure that the graph built from the expert and the one of robot agent possess a same structure at the same time step. 

In the context of cloth folding, we only consider three types of vertices: folding board vertices $\mathcal{V}_b$, cloth vertices $\mathcal{V}_c$, end-effector vertices $\mathcal{V}_e$. Folding board vertices are detected or inferred key points to describe its skeleton during the folding.
Cloth vertices describe the real-time shape (i.e., contour, surface, mesh) during the folding process. Last, the end-effector vertex represents the pose of end-effector bottom which contacts with the folding board.
Structural dissimilarity at a time step $t$ between the human expert graph $\mathcal{G}_t^{E}$ and robot agent graph $\mathcal{G}_t^{R}$ is calculated as our loss function as below:
\begin{multline}
\label{equ:loss}
	\mathcal{L}(\mathcal{G}_t^{E}, \mathcal{G}_t^{R}) =\ \sum_{i \in \mathcal{V}_e, j \in \mathcal{V}_b} w(\mathcal{E}_t^{(i, j)}) \cdot  F(\mathcal{E}_t^{(i, j)} \land f (\mathcal{E}_t^{(i, j)}) ) \cdot \\ \|(v_t^{E, i}-v_t^{E, j})-(v_t^{R, i}-v_t^{R, j})\|
\end{multline}
where $w(\mathcal{E}_t^{(i, j)}) \in \mathbb{R}$ is the weight of the corresponding edge $\mathcal{E}_t^{(i, j)}$ between end-effector vertex set $\mathcal{V}_e$ and folding board vertex set $\mathcal{V}_b$ during imitating clothes folding, and $F$ defined within $\{0, 1\}$ is a Boolean function of the edge $\mathcal{E}_t^{(i, j)}$ and its optical flow function $f (\mathcal{E}_t^{(i, j)})$ to indicate whether $\mathcal{E}_t^{(i, j)}$ is preserved according to current sequence of the refined optical flow.
We regard weight $w(\mathcal{E}_t^{(i, j)})$ as hyperparameters for our framework, and at present we set them with empirical values. 
As for learning optimal values of these weights, we put this important task in our future work.

\section{Methodology} \label{sec:method}
\subsection{Spatio-Temporal Graph (STG) Construction}
\begin{figure}[htbp]
	\centering
	\includegraphics[width=\columnwidth
	]{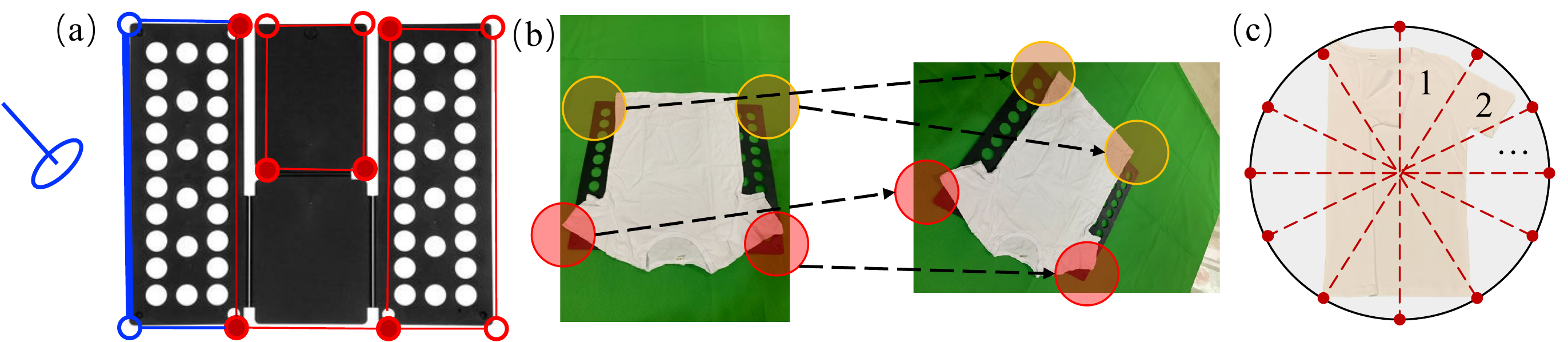}
	\caption{Conceptual representation of STG extractions and clothes state estimation.}
	\label{fig:stg}
\end{figure}

To robustly extract a high-level abstraction for each video frame, we decompose the spatio-temporal graph into three modularized components, namely, 1) \textit{End-effector detection}: the pose of the end-effector must be robustly detected in a real-time manner across the demonstration or imitation workspaces; 2) \textit{Folding board detection}: the graph-structured key points (see Fig. \ref{fig:stg} (a)) will be extracted partially from the original observation and inferred partially from their prior knowledge (i.e., 3D spatial relationships). 3) \textit{Clothing state estimation}: the state of the clothing must be measured after each flipping operation so that we can identify whether the task is solved successfully.

\subsubsection{End-effector Design and Detection}
By analyzing the procedures for each flipping, it is hard to estimate the end-effector's pose at a real-time rate, because sometimes the end-effector could be obscured by the folding board when the end-effector is having contact with the folding board. 
As shown in Fig. \ref{fig:setup} (b), we design and 3D-print modularized components, which includes a bottom circle plate to contact with the folder, a cube with four-sided AR markers to detect its pose, and a long strip to connect the above components. 
Particularly, there is a slope between the upper and lower surfaces of the bottom plate to allow sliding below the folder easily while approaching. In addition, a circular shape design can ensure sufficient and robust contact during lifting, rotating and pushing.
Besides, we calibrate and fix the transformation from the center of the bottom circle plate to each AR marker. 
With this transformation, we can robustly detect the pose of the bottom circle plate over demonstration and imitation workplaces when the folding board obscures the end-effector with close contact in a single camera perspective.

\subsubsection{Folding Board Detection}
We consider folding board detection as a view-invariant correspondence and spatial structure inference problem. 
Although using a folder to assist in folding clothes can avoid the complicated real-time measurement of the highly deformable object, obtaining a precise 3D spatial structure of the folder partially overlapped by the clothes is not an easy task. To alleviate such difficulty, we take the spatial relationships among the key points at the static scene as prior knowledge to infer the entire folder structure. 
Precisely, we first extract four corners (see Fig. \ref{fig:stg}b) with Harris cornet detector at the initial state. 
In case no motions of the folding board occur when the end-effector is approaching the first contact point, we define the edge that will move in the nearest future as an anchor edge.
In contrast, the rest vertices could be calculated according to the initially defined fixed spatial relationships. 
The most challenging folding subtask is the mid-folding where the left and right folding often cover the upper corner points of the mid-section of the folding board.
For this case, we calculate the upper 3D positions based on an average normal of the clothing surface and a fixed spatial transformation defined in the initial state.

With the four corner points, we leverage self-generated visual correspondences to build a powerful dense feature descriptor, which aims at establishing their correlations across the demonstrator and imitator's working place.
To formalize, let $\mathcal{I}$ denote an image with a $h \times w$ resolution, and $\mathcal{D}$ a non-linear point descriptor without modifying its resolution. We are targeting on learning a densely $d$-dimensional descriptor \(\mathcal{D}(\mathcal{I}): \mathbb{R}^{h \times w \times 3} \rightarrow \mathbb{R}^{h \times w \times d}\) to encode the identity for each pixel.
In order to accelerate the training process and optimize computing resource utilization for our dense descriptor,  we employed the backbone of Fully Convolutional Networks \cite{chen2017fast}, 
which could reuse the computed activations for overlapping pixels, from images with arbitrary resolution, and extract the dense descriptor efficiently. 
During the network training process, assume for each flipping we have a video clip \(\mathcal{V}=\{(\mathcal{I}_1, \mathcal{M}_1), \cdots, (\mathcal{I}_N, \mathcal{M}_N)\}\) of $N$ frames, where $\mathcal{M}_n$ corresponds to a universal correspondence model of an image $\mathcal{I}_n$ to provide a mapping from pixels in $\mathcal{I}_n$ to the correspondence coordinates in other frames. 
Only the relationship between different correspondence coordinates in local reference frames are preserved. 
Consequently, we adopt a strategy  of pairwise correspondence coordinate labeling and take a pairwise contrastive loss \cite{chen2017fast} in pixel-level defined as:

\begin{equation}
\begin{split}
&	L(\mathcal{D}(\mathcal{I}_\mathbf{x}), \mathcal{D}(\mathcal{I}^{\prime}_{\mathbf{x}^{\prime}}), \mathcal{M}(\mathcal{I}_{\mathbf{x}}), \mathcal{M}(\mathcal{I}_{\mathbf{x}^{\prime}}^{\prime}))=\\
&	 \left \{  
	\begin{array}{ll}
		\left\|\mathcal{D}\left( \mathcal{I}_{\mathbf{x}}\right)-\mathcal{D}\left( \mathcal{I}^{\prime}_{
		\mathbf{x}^{\prime}
		}\right)\right\|^{2}  & \mathcal{M}(\mathcal{I}_{\mathbf{x}}) = \mathcal{M}(\mathcal{I}^{\prime}_{
		\mathbf{x}^{\prime}
		})
		\\ \max \left(0, \xi-\left\| \mathcal{D}(\mathcal{I}_{\mathbf{x}})- \mathcal{D}(\mathcal{I}^{\prime}_{
		\mathbf{x}^{\prime}}) 
		\right\|\right)^{2} & \text{Otherwise}
	\end{array}
	\right.
\end{split}
\end{equation}
where $\mathcal{D}(\mathcal{I}_{\mathbf{x}})$ and $\mathcal{D}(\mathcal{I}^{\prime}_{\mathbf{x}^{\prime}})$ denote the encoded descriptor for image $\mathcal{I}$ at coordinate $\mathbf{x} = (x, y) $ and image $\mathcal{I}^{\prime}$ at coordinate $\mathbf{x}^\prime = (x^\prime, y^\prime) $, respectively. If the correspondence model of coordinates $\mathbf{x}$ and $\mathbf{x}^\prime$ can be mapped to the same 3D point, we regard them as a positive pixel pair to minimize the distance in the feature space built with the correspondence descriptor. 
Otherwise, the contrastive loss will split them at least $\xi$ margin away. 
We automate the collection of image data of the robotic agent's working space from different perspectives by using a hand-eye calibrated RGB-D camera, which is installed on the end-effector. 
Then the camera will move along with random trajectories that cover different perspectives of the workplace.
The camera pose is estimated via hand-eye calibration using the robot's forward kinematics model. Based on the aligned depth image and camera's intrinsic parameters, we can reconstruct the spatial configuration of the workspace across different perspectives.

\begin{figure*}[h]
	\centering
	\includegraphics[width=2\columnwidth]{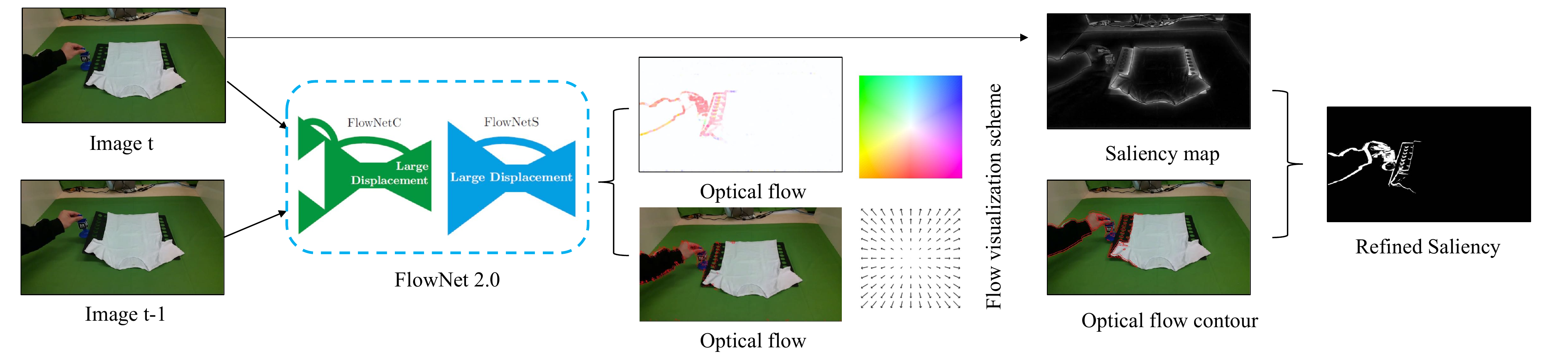}
	\caption{The flow chart of the refined optical flow-based motion saliency map. Given an image pair at time step $t$ and $t-1$, $\mathcal{I}_{t}$ and $\mathcal{I}_{t-1}$, an optical flow $\mathcal{F}_{t}$ can be detected with FlowNet 2.0 and then be refined based on the motion contour $\mathcal{C}_{t}$ extracted from a static motion saliency $\mathcal{S}_{t}$.
	}
	\label{fig:motion}
\end{figure*}

\subsubsection{Clothing State Estimation}
Clothing state identification will be executed after each folding task. To robustly estimate the state from multiple perspectives, we build a simple but efficient classifier by combining radial region-based features and  view-invariant moments features. 
As shown in Fig. \ref{fig:stg}c, we divide the mask of the extracted clothing into 12 radial bins after computing the centroid as below:
\begin{equation}
	C_{m}=\left( \frac{\sum_{i=1}^{n} x_{i}}{n}, \frac{\sum_{i=1}^{n} y_{i}}{n} \right)
\end{equation}
For each radial area, we compute a feature vector formed by averaging critical points, area, eccentricity, perimeter and orientation similarly defined in \cite{sargano2016human}. Based on Hu-Moments invariant features \cite{hu1962visual}, the moment \((p, q)\) of an image \(f(x, y)\) of size \(M \times N\) is defined as: \(m_{p, q}=\sum_{x=0}^{M-1} \sum_{y=0}^{N-1} f(x, y) x^{p} y^{q}\), where \(p\) and \(q\) are the order of \(x\) and \(y\), respectively. We compute the clothing moment similarly, except that \(x\) and \(y\) are displaced by the mean values as follows:
\begin{equation}
\mu_{p, q}=\sum_{x=0}^{M-1} \sum_{y=0}^{N-1} f(x, y)\left(x-\bar{x}\right)^{p}\left(y-\bar{y}\right)^{q}	
\end{equation}
where $\bar{x}=\frac{m_{10}}{m_{00}}$ and $\bar{y}=\frac{m_{01}}{m_{00}}$

By applying normalization, clothing moments are defined as follows:
\begin{equation}
	\eta_{\mathrm{p}, \mathrm{q}}=\frac{\mu_{\mathrm{p}, \mathrm{q}}}{\mu 00^{\gamma}}, \gamma=\frac{\mathrm{p}+\mathrm{q}+2}{2}, \mathrm{p}+\mathrm{q}=2,3, \ldots
\end{equation}
Finally, based on these clothing moments, 7-dimensional Hu-moments could be calculated. By concatenating two group features, a Support Vector Machine (SVM) is implemented to identify whether a clothing state is a correct one after the corresponding folding.

\subsection{Refined Optical Flow-based STG}
After completing the STG construction, we propose extracting the refined optical flow from the object saliency map to determine which edge must be preserved during the policy learning process.
We anticipate that the optical flow field should be sufficiently identifiable in prominent regions. To this end, we will compare the flow field with a real-time static saliency map which would have enclosed the calculated optical flow in the identical regions. 
The former may be determined using an optical flow technique. The latter is not immediately accessible because it is not noticed. Nonetheless, it is predictable using a refined optical flow technique. This is precisely the novelty of utilizing refined flow to denote real-time motion saliency to identify the presence of STG edges.

Our overall framework is illustrated in Fig. \ref{fig:motion}. Given two adjacent frames $\mathcal{I}_t$ and $\mathcal{I}_{t-1}$ at time step $t$ and $t-1$ in the video clippings, we first calculate its real-time optical flow $\mathcal{F}_t$ with FlowNet 2.0 \cite{ilg2017flownet}. However, the motion boundary generated with FlowNet 2.0 lacks details in few small but important regions. Therefore, we compute another object saliency map with Spectral Residual (SR) algorithm \cite{hou2007saliency} $\mathcal{S}_t$ to paint the raw optical flow in order to yield a refined optical flow to indicate the motion saliency in a real-time manner. For the motion contour, we could directly employ a threshold on the norm of the gradient of the velocity vectors. However, this method will yield noisy contours. Instead, we choose to perform the classical Canny \cite{canny1986computational} edge detection method on the transformed image with an HSV color scheme. Then the region enclosed by motion contour $\mathcal{C}_t$ generated with respect to the raw optical flow $\mathcal{F}_t$ is regarded as an inpainted mask $\mathcal{C}_t$ to cut the static saliency map $\mathcal{S}_t$. Finally, a threshold is performed to yield the final refined flow $\Re_t$.

The refined optical flow, i.e., the object saliency map $\mathcal{S}_t$ located inside the motion contour $\mathcal{C}_t$ enclosed by the optical flow $\mathcal{F}_t$, is computed over the inpainted mask indicated by the motion contour. From the refined flow, a motion saliency map is expected to be defined within $[0, 1]$ as follows:
\begin{equation}
	\forall \mathbf{x} \in \mathcal{C}_t, 
	\quad \Re(\mathbf{x}) = 1 - \exp(-\lambda \| \mathcal{R}_t(\mathbf{x})  \|_{2})
\end{equation}
where $\Re(\mathbf{x}) = 0$ for $\mathbf{x} \notin \mathcal{C}_t$ and $\lambda$ regulates the refined optical flow score. Therefore, $\Re(\mathbf{x})$ output a non-zero flow score highlighting the moving elements, while $\lambda$ can be regarded as a trade-off between robustness to noise and ability to highlight tiny but still salient motions. Specifically, we can set a high $\lambda$ to produce a binary map to explicitly present motion segmentation, as shown in Fig. \ref{fig:motion}$(\mathrm{e})$. To this end, we introduce a parameter $\varepsilon$ to segment the refined flow as below:
\begin{equation}
	\| \mathcal{R}_t(\mathbf{x})  \|_{2} \geqslant -\frac{\ln (1-\varepsilon)}{\lambda}
\end{equation}
In this case, the refined flow magnitude at an image location $\mathbf{x}$ larger than $\frac{\ln 2}{\lambda}$ will be segmented out if $\varepsilon$ is set to $0.5$. Therefore, the introduction of $\lambda$ is able to add relative flexibility into this workflow built for refined optical flow.

\subsection{Motion Policy with Refined Flow-based STG}
We cast motion policy learning using a refined flow-based STG as a reinforcement learning problem. With the optimized policy, we are aiming at teaching the robot to learn cloth folding from a single demonstration video of human experts. Formally, for each time step $t$ of the cloth folding task, the $\theta$-parameterized policy ${\pi}_{\theta}\left(\mathbf{a}_{t} \mid \mathbf{s}_{t}\right)$ identifies a probability distribution over actions $\mathbf{a}_{t}$ constrained by current system state \(\mathbf{s}_{t}\). Let \(\tau=\left(\mathbf{s}_{1}, \mathbf{a}_{1}, \ldots, \mathbf{s}_{T}, \mathbf{a}_{T}\right)\) denote a trajectory of states and actions of cloth folding. With a loss function $l(\mathbf{s}_{t}, \mathbf{a}_{t}) $, the entire trajectory cost can be formulated as:
\begin{equation}
	J(\theta)=\mathbb{E}_{\pi}[l(\tau)]=\int l(\tau) \pi(\tau) d \tau
\end{equation}  
where $\pi(\tau)$ is the distribution of policy trajectory under the system dynamics $\pi\left(\mathbf{s}_{t+1} \mid \mathbf{s}_{t}, \mathbf{a}_{t}\right)$ defined as below:
\begin{equation}
	\pi(\tau)=\pi\left(\mathbf{s}_{1}\right) \prod_{t=1}^{T} \pi\left(\mathbf{s}_{t+1} \mid \mathbf{s}_{t}, \mathbf{a}_{t}\right) \pi\left(\mathbf{a}_{t} \mid \mathbf{s}_{t}\right)
\end{equation}
 The action $\mathbf{a}_t$ is defined as relative pose alteration $\Delta\mathbf{p}_t = \mathbf{p}_t - \mathbf{p}_{t-1}$ of the robot end-effector, which is composed of a 7-dimensional vector $ (\Delta x, \Delta y, \Delta z, \Delta q_1, \Delta q_2, \Delta q_3, \Delta q_4)$, where $\Delta x, \Delta y, \Delta z$ represent the position's shift and $\Delta q_1, \Delta q_2, \Delta q_3, \Delta q_4)$ denotes the orientation's shift. The state $\mathbf{s}_t$ is defined as a vector consisting of end-effector's 3D positions, robot joint angles and graph spatial configurations of the workspace.
  For $n$ detected anchor points, a \(3 + n_{\text {joints }} + n_{\text {anchor\_points }} * 3 + d_{\text {anchor\_points }}\) 
  dimensional state space is formed with this vectorization, where \(n_{\text {joints }}\) denotes the dimensions of the robot joints, 
  $n_{\text {anchor\_points }}$ is the anchor points detected from the active section of the folding board, and $d_{\text {anchor\_points }}$ indicates the distances between the end-effector and different anchor points. In our case, the distance between different anchor points are ignored because their spatial relationships are relatively fixed on the folding board. While for imitation learning with multiple objects, it is still needed to preserve.
  

To take the advantages of model-free and model-based reinforcement learning approaches at the same time, a state-of-the-art technique called PILQR \cite{chebotar2017combining} is applied to minimize the entire trajectory cost defined in Eq. (\ref{equ:loss}). This approach optimizes TVLG policies by combining rapid model-based updates via iterative linear-Gaussian model fitting and improved model-free updates in the $\text{PI}^2$ framework. Consequently, 
it is capable of combining model-based learning's efficiency with the generality of model-free updates with the aim at solving complicated continuous control problems, which are infeasible to perform solely using either linear-Gaussian models or $\text{PI}^2$ alone. Besides, it is able to maintain orders of magnitude more efficient than conventional model-free RL.
Finally, a time-dependent policy is learned as below:
\begin{equation}
\pi_{t}\left(\mathbf{a}_{t} \mid \mathbf{s}_{t} ; \theta\right)=\mathcal{N}\left(\mathbf{K}_{t} \mathbf{s}_{t}+\mathbf{k}_{t}, \boldsymbol{\Sigma}_{t}\right)
\end{equation}
During the training, the control gains are optimized via mixed model-based and model-free updates. Thus, the unknown temporal system dynamics \(\pi\left(\mathbf{s}_{t+1} \mid \mathbf{s}_{t}, \mathbf{a}_{t}\right)\) is learned.
 


\begin{figure}[htbp]
	\centering
	\includegraphics[width=\columnwidth
	]{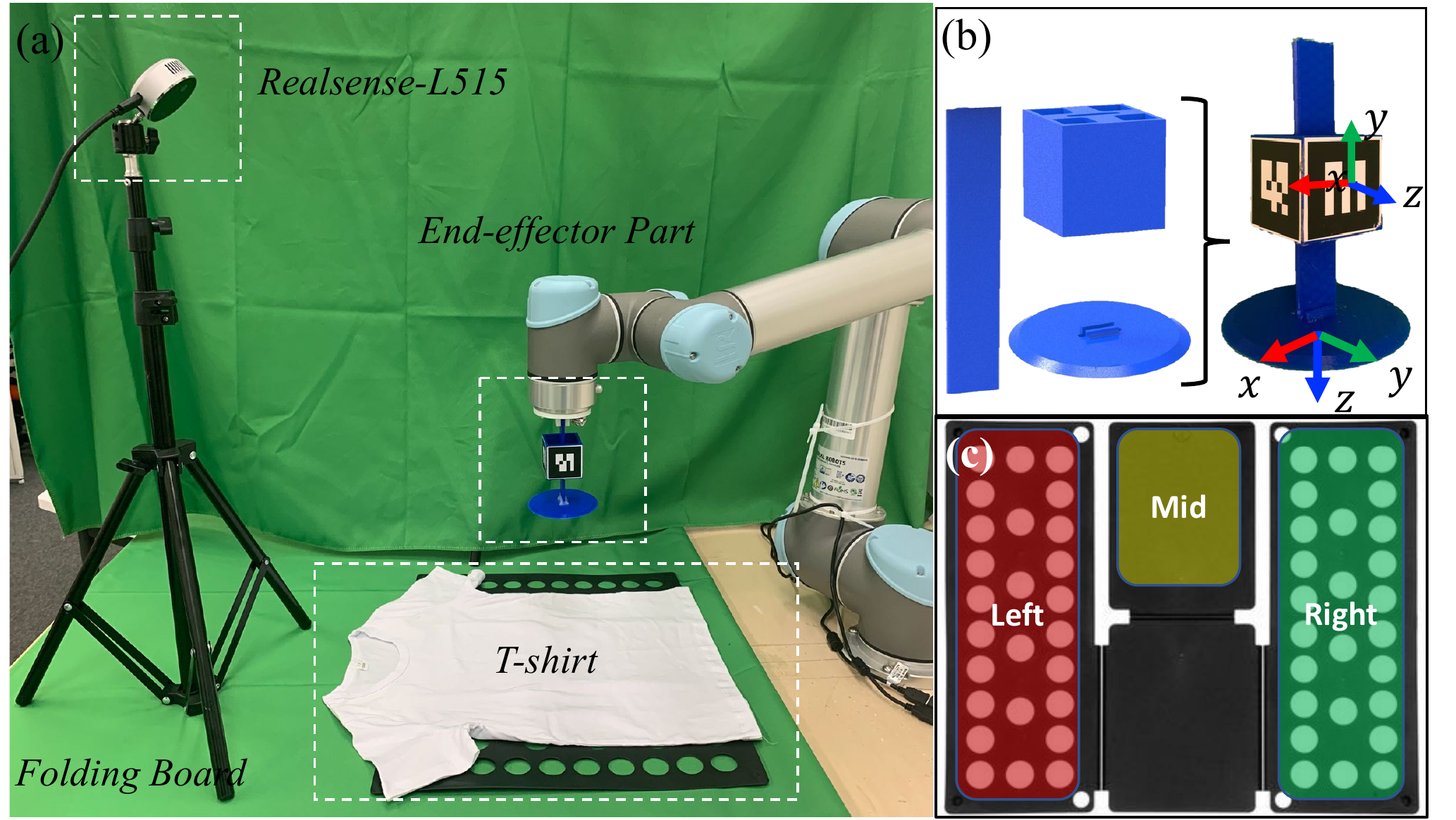}
	\caption{(a) shows the experimental setup for learning cloth folding task using an assistive folding board; (b) presents the designed end-effector for a robust pose estimation; (c) gives the top view of a folding board and different color represents different folding section.}
	\label{fig:setup}
\end{figure}



\section{Results} \label{sec:result}
In this section, the experimental setup for learning cloth folding is described first, and afterward, the learned policy with our approach is compared with the one with a basic STG and other existing approaches. 
 Three tasks performed on a T-shirt are considered to imitate: \textbf{left folding}, \textbf{right folding} and \textbf{mid-folding}. For each folding, 
 we define four procedures, namely, \textit{approaching}: the end-effector is approaching the first contact position to insert the bottom circle board between the left side and workbench; \textit{lifting}: the end-effector is lifting up to obtain enough space for rotating; \textit{rotating}: the end-effector is rotated to align with the folding board plane; \textit{pushing}: the end-effector is pushing the corresponding folder section with a certain speed to finish the task. To evaluate the generalizability, we also extend the approach to the shorts folding task.
 
%

\subsection{Experiment Setup}
To validate the effectiveness of the policy learned via refined optical flow-based STGs from a single demonstration video in a robotic cloth folding task using a folding board, a UR-5 robot is utilized to perform the different folding tasks with a well-designed end-effector. As shown in Fig. \ref{fig:setup}, in order to robustly detect the real-time pose of the end-effector, we design and 3D-print modularized components as shown in Fig. \ref{fig:setup}, which includes a bottom circle plate to contact with the folder, a cube with four-sided AR markers to detect its pose, and a long strip to connect the above components. We calibrate and fix the transformation from the center of the bottom circle plate to each AR marker. With this transformation, we can robustly detect the pose of the bottom circle plate over demonstration and imitation workplaces when the folding board obscures the end-effector with close contact in a single camera perspective. 
The entire experimental setup is shown in Fig. \ref{fig:setup}, where we use an RGB-D camera (RealSense L-515) to observe the cloth folding manipulation process using a folding board (see Fig. $\ref{fig:setup} (\mathrm{c})$, and the clothing (T-shirt or shorts) is selected to evaluate the generated folding policy. Note that each piece of clothing is laid out on the board smoothly and flat at the initial state.

\subsection{Comparison with Existing Methods}
Time-contrastive networks (TCN) \cite{sermanet2018time} --- a self-supervised approach for robotic imitation learning --- and a pure STG are selected to compare against our learning approach using the refined optical flow based-STG. TCN trains a viewpoint-invariant  representation in an embedding space ($g:\mathcal{X} \rightarrow \mathcal{Z}$) with a reward function defined based on the squared Euclidean distance and a Huber-style loss, which is formulated as below:
\begin{equation}
	R\left(\mathbf{z}_{t}^{E}, \mathbf{z}_{t}^{R}\right)=-\alpha\left\|\mathbf{z}_{t}^{E}-\mathbf{z}_{t}^{R}\right\|_{2}^{2}-\beta \sqrt{\gamma+\left\|\mathbf{z}_{t}^{E}-\mathbf{z}_{t}^{R}\right\|_{2}^{2}}
\end{equation}
where $\mathbf{z}_{t}^{E}$ and $\mathbf{z}_{t}^{R}$ is the TCN embedding calculated from the demonstration video sequences of human experts and the imitation video sequence of robot agents, respectively. Here, \(\alpha\) and \(\beta\) are weighting parameters to control policy updates magnitude and task execution precision. 
In our experiment, we implement the exact same network architecture designed in \cite{sermanet2018time}, where there are two convolutional layers, a spatial softmax layer and a fully-connected output layer linked to the Inception model pre-trained using ImageNet up to the Mixed 5d level. Finally, it transforms the input image from the original space into a 32-dimensional embedding space.
Sixteen video recordings, eight human expert demonstrations, and eight robot imitation executions for each folding task are collected to train a corresponding TCN representation. In contrast, one demonstration is provided for learning the folding policy with our refined optical flow-based STG in the same system and environment configurations. 
During training session of TCN models, we set $\alpha=0.8$, $\beta=0.001$, $\gamma = 10^{-5}$ respectively.
Though only one demonstration video is needed for our approach training, policy searching with the refined optical flow-based STG requires a view-invariant dense correspondence descriptor and creating a refined optical flow. 
Thus, the same size of data are collected to train the TCN baseline for a fair comparison.
%
To compare the proposed refined spatiotemporal graph encoding with raw spatiotemporal encoding and previous pixel encoding \cite{sermanet2018time}, we conduct experiments to test the effectiveness of our method against multiple viewpoints, as well as its generalizability across clothing of various categories and textures. We also assess the method's robustness to the changes in initial object spatial configurations.

\begin{figure*}[h]
	\centering
	\includegraphics[width=2\columnwidth]{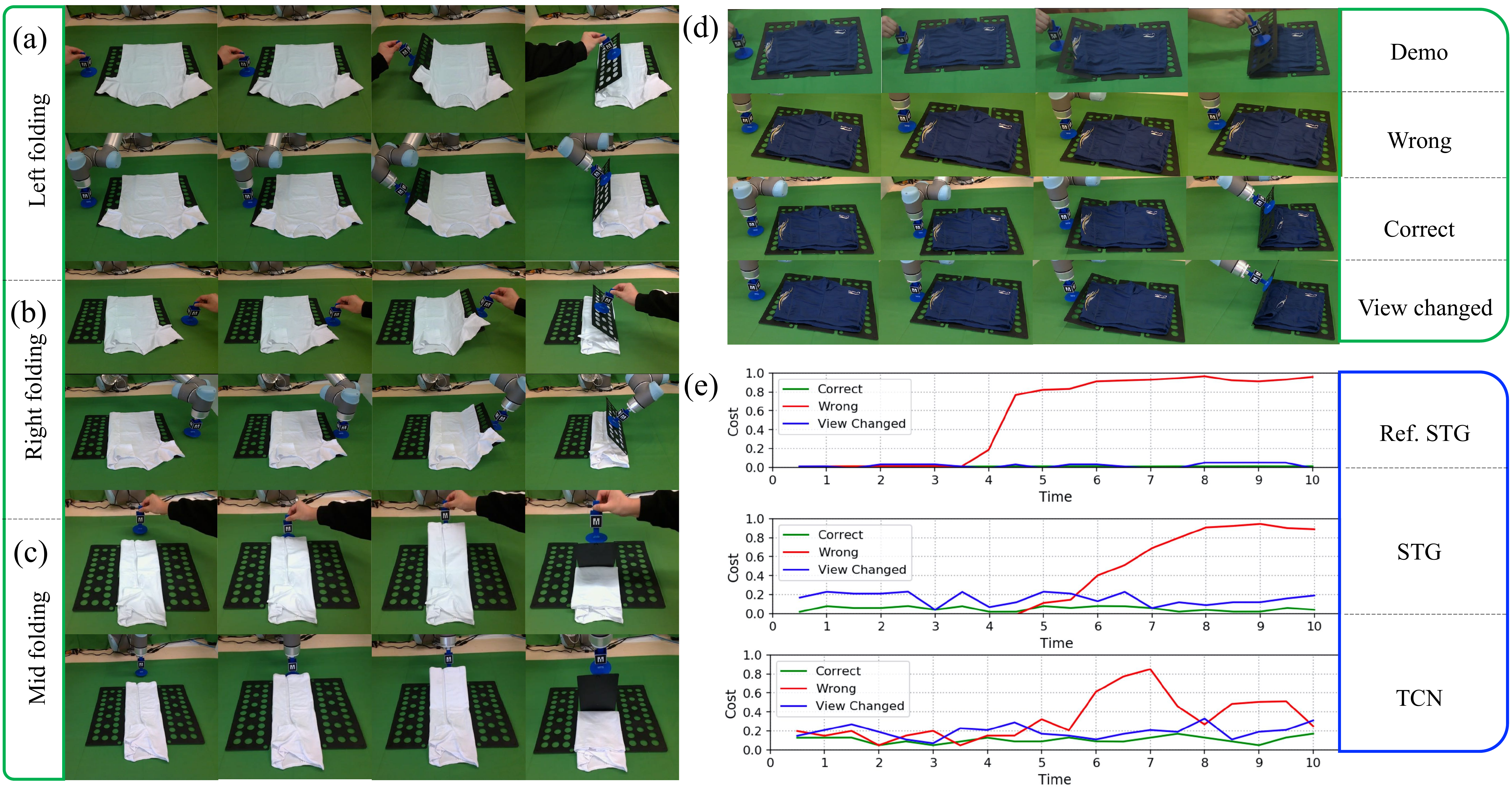}
	\caption{Performance of the designed folding task. (a)-(c) present the	successful folding examples for different folding tasks, respectively. (d) shows different robot imitation videos to compare the imitation efficiency.
	(e) shows the corresponding reward curves of different approaches on above imitation videos.}
	\label{fig:result}
\end{figure*}

Fig. $\ref{fig:result} (\mathrm{e})$ depicts our method's reward curves as well as the pure STG and TCN baseline for different robot imitation videos, indicating how effectively the robot learns the demonstrated skill.
The vertical axis indicates the imitation cost, while the horizontal represents the imitation time.
Despite viewpoint changed in the 3rd row, the proposed cost function for refined spatiotemporal graphs defined in Equ. (\ref{equ:loss}) detects all valid positive imitations and accurately reports the incorrect imitation period in the wrong curve. 
The baseline TCN cost curves, on the other hand, are non-discriminative. 
Lastly, though the STG cost curve shows a relatively higher than TCN curve, it can not compete with our method. 
Cost curves with high discrimination are essential for successful policy learning, and details will be discussed in the following.

 


\begin{table*}[htbp]
\caption{Success rate of left folding, right folding and mid-folding tasks on a T-shirt.}
\label{tab:data_sum}
\centering
\subfloat
{
        \begin{tabular}{lccc}
            \toprule
            Process & Ref. STG  & STG & TCN \\
            \midrule 
Approaching & \(8 / 8\) & \(8 / 8\) & \(3 / 8\) \\
Lifting     & \(8 / 8\) & \(7 / 8\) & \(2 / 3\) \\
 Rotating    & \(8 / 8\) & \(6 / 7\) & \(2 / 2\) \\
 Pushing     & \(7 / 8\) & \(5 / 6\) & \(1 / 2\) \\
 Summary     & \(7 / 8\) & \(5 / 8\) & \(1 / 8\) \\
            \bottomrule
        \end{tabular}

}
\qquad
\subfloat
{
        \begin{tabular}{lccc}
            \toprule
            Process & Ref. STG  & STG & TCN \\            
            \midrule       
 Approaching & \(8 / 8\) & \(8 / 8\) & \(4 / 8\) \\
 Lifting     & \(8 / 8\) & \(8 / 8\) & \(3 / 4\) \\
 Rotating    & \(8 / 8\) & \(7 / 8\) & \(3 / 3\) \\
 Pushing     & \(8 / 8\) & \(5 / 7\) & \(1 / 3\) \\
 Summary     & \(8 / 8\) & \(5 / 8\) & \(1 / 8\) \\
            \bottomrule
        \end{tabular}
}
\qquad
\subfloat
{
        \begin{tabular}{lccc}
            \toprule
            Process & Ref. STG  & STG & TCN \\            
            \midrule       
 Approaching & \(8 / 8\) & \(8 / 8\) & \(3 / 8\) \\
 Lifting     & \(7 / 8\) & \(7 / 8\) & \(2 / 3\) \\
 Rotating    & \(7 / 7\) & \(6 / 7\) & \(1 / 2\) \\
 Pushing     & \(6 / 7\) & \(4 / 6\) & \(0 / 1\) \\
 Summary     & \(6 / 8\) & \(4 / 8\) & \(0 / 8\) \\
            \bottomrule
        \end{tabular}
}
\vspace{0cm}
\end{table*}

\subsection{Discussion}
In this task, first, the clothing is laid out in the center of the folding board smoothly and flat. Second, according to the common practice for clothing folding, the robot needs to go through four different processes. Namely, \romannum{1}) Approaching \romannum{2}) Lifting, \romannum{3}) Rotating, and \romannum{4}) Pushing. For each folding, the robot will execute the abovementioned four common processes with slight motion changes.

In order to measure the performance of the refined STG approach, we set a home position setting above the folding board with a certain distance (50 cm). For each method, we run the policy eight times, starting from the home position. We consider the task success only when the four sub-processes are all solved because these four sub-processes are executed sequentially. Any unsuccessful sub-process will lead to the failure of the entire task. The detailed success rate of each process for the different folding tasks is reported in Table \ref{tab:data_sum}. The robot successfully solves the left folding for seven runs out of 8 with our proposed approach, demonstrating a solid ability to handle complexity and dynamics. STG gets a success rate of 5 runs out of 8, which is acceptable compared with the TCN-based approach. 
TCN failed in almost all runs at the first sub-process. At the same time, STG performs better at the beginning of each process since graph-structured state representation can capture and handle spatiotemporal dynamics.
However, it performs poorly when handling the rest folding process such as \textit{pushing} because only the part of the related motion is vital for policy-shaping instead of the entire structure of the STG. With refined optical flow, our method can help reinforcement learning algorithms optimize the policy efficiently, focusing on system dynamics caused by the detected core motions in every policy update.

Right folding is similar to left folding because they are mirror operations in spatial configurations. Therefore, the success rate shares the same pattern with the left folding task. Specifically, our method completed this task for all runs out of 8, and one of the cases is shown in Fig. \ref{fig:result}.
On the other hand, the mid-folding is a challenging task of imitating from the demonstration videos. As shown in Fig. \ref{fig:result}, the viewpoint of the camera is in the front of the folding board with an article of clothing after two folds.
Therefore, the graph difference caused by motion changes is harder to detect compared to the other abovementioned folding tasks, which is aligned with the success rate shown in the Tab. \ref{tab:data_sum}. 
From the table, we can tell that all testing approaches perform worse at mid-folding compared to the other folding subtasks. 
Specifically, TCN failed all the runs finally. On the other hand, STG only has a half success rate for four runs out of 8. However, our method sill can successfully complete six runs out of 8 in total. Generally, TCN stands for the policy searching purely based on the states laying in the embedding space generated from pure image inputs, which handles complex spatiotemporal dynamics poorly, while the STG-based approach handles such dynamics better. However, it lacks an attentional motion mechanism that can focus on the core structure caused by the significant motions and help accelerate policy learning updates.

\subsection{Limitations}
We attempt to solve clothing folding in a perspective of visual imitation using a refined optical flow-based spatiotemporal graph structure as its input. 
Consequently, constructing a precise refined flow-based graph is crucial because a precise graph will result in a precise system state representation, thus leading to the policy updating and optimizing in an explicit and accurate manner. 
However, cloth folding highly relates to occlusion, and our current visual corresponding method can not deal with fully occluded objects. Furthermore, secondly, training such dense corresponding descriptors is time-consuming. On the other hand, how to imitate a high-speed motion (i.e., the robot in pushing sub-process requires a relatively high-speed motion) remains an open question, and little research relates to this topic at present. 
To fully solve this problem, multi-perspective or active vision might help collect more valuable observations from a vantage viewpoint, thus accelerating the policy searching process.
Another interesting topic for future work would be encoding prior knowledge such as physical models into imitation learning policy learning.


\section{Conclusion} \label{sec:con}
In this paper, we proposed a solution for cloth folding by learning from video demonstration using a refined optical flow-based STG as the input. With a dense corresponding descriptor, we identify the intended pixel between different video frames to construct a basic STG. Subsequently, we combine optical flow and static motion saliency map, aiming at refined STG with attentional motion mechanism. Experimental results on a real robotic platform have validated the effectiveness and robustness of the proposed approach. 
As future research, we plan to study how to optimize and update the policy learning from cloth folding with the assistance of a folding board to the task without the folding board.

\bibliographystyle{IEEEtran}
\bibliography{root}
\end{document}